\documentclass[journal]{IEEEtran}
\usepackage{amsmath,epsfig,algorithm,algorithmic, amssymb, balance, subfigure, makecell, array,cite,subfigure,enumerate,multirow}
 
\newcolumntype{x}[1]{
>{\centering\hspace{0pt}}p{#1}}

\begin{document}

\title{An Ensemble-based System for Microaneurysm Detection and Diabetic Retinopathy Grading}

\author{B\'alint~Antal,~\IEEEmembership{Student Member,~IEEE,}
        and~Andr\'as~Hajdu,~\IEEEmembership{Member,~IEEE}
\thanks{B\'alint~Antal and Andr\'as~Hajdu are with the University of Debrecen, Faculty of Informatics, POB 12, 4010, Debrecen, Hungary. E-mail: \{antal.balint, hajdu.andras\}@inf.unideb.hu. The corresponding author is B\'alint~Antal. Phone: +36 52 512-900/62830, Fax: +36 52 512-900/62822.}}

\markboth{Transactions on Biomedical Engineering}%
{An Ensemble-based System for Microaneurysm Detection and Diabetic Retinopathy Grading}

\maketitle

\begin{abstract}

Reliable microaneurysm detection in digital fundus images is still an open issue in medical image processing. We propose an ensemble-based framework to improve microaneurysm detection. Unlike the well-known approach of considering the output of multiple classifiers, we propose a combination of internal components of microaneurysm detectors, namely preprocessing methods and candidate extractors. We have evaluated our approach for microaneurysm detection in an online competition, where this algorithm is currently ranked as first and also on two other databases. Since microaneurysm detection is decisive in diabetic retinopathy grading, we also tested the proposed method for this task on the publicly available Messidor database, where a promising AUC 0.90 with 0.01 uncertainty is achieved in a 'DR/non-DR'-type classification based on the presence or absence of the microaneurysms.

\end{abstract}


\section{Introduction}
\IEEEPARstart{D}{iabetic} retinopathy (DR) is a serious eye disease that originates from diabetes mellitus and is the most common cause of blindness in the developed countries. Early treatment can prevent patients to become affected from this condition or at least the progression of DR can be slowed down. Thus, mass screening of patients suffering from diabetes is highly desired, but manual grading is slow and resource demanding. Therefore, several efforts have been made to establish reliable computer-aided screening systems based on color fundus images \cite{system}. The promising results reported by Fleming et al. \cite{largescale} and Jelinek et al. \cite{Jelinek2006} indicates that automatic DR screening systems are getting closer to be used in clinical settings. 

A key feature to recognize DR is to detect microaneurysms (MAs) in the fundus of the eye. The importance of handling MAs are two-fold. First, they are normally the earliest sign of DR, hence their timely and precise detection is essential. On the other hand, the grading performance of computer-aided DR screening systems highly depends on MA detection \cite{Jelinek2006} \cite{early}. In this paper, we propose a microaneurysm detector which provides remarkable results from both aspects.   

One way to ensure high reliability and raise accuracy in a detector is to consider ensemble-based systems, which have been proven to be efficient in several fields. However, the usual ensemble techniques aim to combine class labels or real values which cannot be adopted in our case. In MA detection, detectors provide spatial coordinates as centers of potential MA candidates. The use of well-known ensemble techniques would require a classification of each pixel, which can be misleading in our context, since different algorithms extract MAs with different approaches and the MA centers may not coincide exactly. To overcome this difficulty, we gather close MA candidates of the individual detectors and apply a voting scheme on them. 

\begin{figure}[htb]
	\centering
		\includegraphics[width=0.5\linewidth]{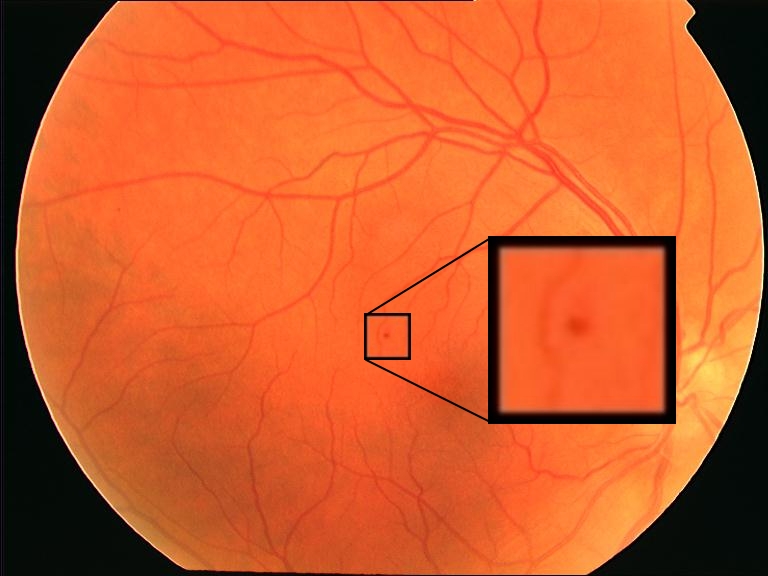}
	\caption{Sample digital fundus image with a microaneurysm.}
	\label{fig:ma_large_zoomed}
\end{figure}

In \cite{combining}, Niemeijer et al. showed that the fusion of the results of the several MA detectors lead to an increased average sensitivity measured at seven predefined false positive rates. In this paper, we propose a framework to build MA detector ensembles based on the combination of the internal components of the detectors not only on their output as in \cite{combining}. Some of our earlier research on combining MA detectors did not provide reassuring results \cite{sofa}. To increase the accuracy of such ensembles, we must identify the weak points of MA detection. The first difficulty originates from the shape characteristics of MAs. They appear as small circular dark spots on the surface of the retina (see Figure \ref{fig:ma_large_zoomed}), which can be hard to distinguish from fragments of the vascular system or from certain eye features. Most MA detectors tackle this problem in the following way: first, the green channel of the fundus image is extracted and preprocessed to enhance MA like characteristics. Then, in a coarse level step (which will be referred as candidate extraction in the rest of the paper), all MA-like objects are detected in the image. Finally, a fine level algorithm (usually a supervised classifier) removes the potentially false detections based on some assumptions about MAs. Our former investigations showed that the low sensitivity of MA detectors originates from the candidate extractor part \cite{pr}. However, we could increase the sensitivity by applying proper preprocessing methods before candidate extraction. This technique causes a slight increment in the number of false positives, but it can be decreased by classification or voting. 

In this paper, we propose an effective microaneurysm detector based on the combination of preprocessing methods and candidate extractors. We provide an ensemble creation framework to select the best combination. An exhaustive quantitative analysis is also given to prove the superiority of our approach over individual algorithms. We also investigate the grading performance of our method, which is proven to be competitive with other screening systems.

The rest of the paper is organized as follows: the selected preprocessing methods and candidate extractors are presented in section \ref{sec:pre} and \ref{sec:ces}, respectively. The details of the proposed ensemble creation framework is discussed in section \ref{sec:com}. We present our evaluation methodology in section \ref{sec:methodology}. In section \ref{sec:res}, we summarize our experimental results. A detailed discussion is given in section \ref{sec:dis} to address several issues. Finally, we draw conclusions in section \ref{sec:con}.  

\section{Preprocessing methods}
\label{sec:pre}

In this section, we present the selected preprocessing methods, which we consider to be applied before executing MA candidate extraction. The selection of the preprocessing method and candidate extractor components for this framework is a challenging task. Comparison of preprocessing methods dedicated to microaneurysm detection has not been published yet. Since preprocessing methods need to be highly interchangeable, we must select algorithms which can be used before any candidate extractor and do not change the characteristics of the original images (unlike e.g. shade correction \cite{youssif}). We also found some techniques to generate too noisy images for MA detection (histogram equalization \cite{youssif}, adaptive histogram equalization \cite{youssif} or color normalization \cite{youssif}). Thus, we have selected methods which are well-known in medical image processing and preserve image characteristics. Naturally, the proposed system can be improved in the future with adding new methods. A summary on the key differences of the algorithms is given in Table \ref{tab:pps}.

\subsection{Walter-Klein contrast enhancement \cite{ce}}
\label{sec:ce}

This preprocessing method aims to enhance the contrast of fundus images by applying a gray level transformation using the following operator:
\[
f' = \begin{cases} 
		\dfrac{\dfrac{1}{2} \left(f'_{max} - f'_{min}\right)}{\left(\mu - f_{min}\right)^r} \cdot \left(f - f_{min}\right)^r + f'_{min}, & f \leq \mu, \\ 
		\dfrac{- \dfrac{1}{2} \left(f'_{max} - f'_{min}\right)}{\left(\mu - f_{max}\right)^r} \cdot \left(f - f_{max}\right)^r + f'_{max}, & f \geq \mu,
	  \end{cases}
\]
where $\{f_{min},\,\dots,\,f_{max}\},\,\{f'_{min},\,\dots,\,f'_{max}\}$ are the intensity levels of the original and the enhanced image, respectively, $\mu$ is the mean value of the original grayscale image and $r \in \mathbb{R}$ is a transition parameter. 

\subsection{Contrast limited adaptive histogram equalization \cite{clahe}}
\label{sec:clahe}

Contrast limited adaptive histogram equalization (CLAHE) is a popular technique in biomedical image processing, since it is very effective in making the usually interesting salient parts more visible. The image is split into disjoint regions, and in each region a local histogram equalization is applied. Then, the boundaries between the regions are eliminated with a bilinear interpolation. 

\subsection{Vessel removal and extrapolation \cite{cvpr}}
\label{sec:vre}

We investigate the effect of processing images with the complete vessel system being removed based on the idea proposed in \cite{cvpr}. We extrapolate the missing parts to fill in the holes caused by the removal using the inpainting algorithm presented in \cite{inpaint}. MAs appearing near vessels become more easily detectable in this way.

\subsection{Illumination equalization \cite{youssif}}
\label{sec:ie}

This preprocessing method aims to reduce the vignetting effect caused by uneven illumination of retinal images. Each pixel intensity is set according to the following formula:
\[
	f' = f + \mu_{d} - \mu_{l},
\]
where $f, f'$ are the original and the new pixel intensity values, respectively, $\mu_{d}$ is the desired average intensity and $\mu_{l}$ is the local average intensity. MAs appearing on the border of the retina are enhanced by this step.

\subsection{No preprocessing}
\label{sec:nopp}

We also consider the results of the candidate extractors obtained for the original images without any preprocessing. That is, we formally consider a "No preprocessing" operation, as well.

\begin{table}
\centering
\caption{Summary of the key differences of the preprocessing methods.}
\label{tab:pps}  
\begin{tabular}{|l|l|l|}
\hline
Algorithm & Aim & Method\\
\hline
\multirow{2}{*}{Walter-Klein} & \multirow{2}{80pt}{contrast enhancement} & \multirow{2}{70pt}{gray level transformation}\\
&&\\
\hline
\multirow{2}{*}{CLAHE} & \multirow{2}{80pt}{salient object enhancement} & \multirow{2}{70pt}{local histogram equalization}\\
&&\\
\hline
\multirow{2}{*}{Vessel Removal} & \multirow{2}{80pt}{MA enhancement near vessels} & \multirow{2}{70pt}{vessel removal and inpainting}\\
&&\\
\hline
\multirow{2}{*}{Illumination eq.} & \multirow{2}{80pt}{MA enhancement at the border of the ROI} & \multirow{2}{*}{vignette correction}\\
&&\\
\hline
\end{tabular}
\end{table}


\section{Microaneurysm candidate extractors}
\label{sec:ces}

Candidate extraction is a process which aims to spot any objects in the image showing MA-like characteristics. Individual MA detectors consider different principles to extract MA candidates. In this section, we provide a brief overview of the candidate extractors involved in our analysis. Again, just as for preprocessing methods, adding new MA candidate extractors may lead to further improvement in the future. A summary on the key differences of the candidate extractor algorithms and their performance measured in the ROC training dataset \cite{roc} are shown in Table \ref{tab:ces}. 

\subsection{Walter et al. \cite{walter}}
\label{sec:walter}

Candidate extraction is accomplished by grayscale diameter closing. That is, this method aims to find all sufficiently small dark patterns on the green channel. Finally, a double threshold is applied.

\subsection{Spencer et al. \cite{spencer}}
\label{sec:spencer}

From the input fundus image, the vascular map is extracted by applying twelve morphological top-hat transformations with twelve rotated linear structuring elements (with a radial resolution $15\,^{\circ}$). Then, the vascular map is subtracted from the input image, which is followed by the application of a Gaussian matched filter. The resulting image is then binarized with a fixed threshold. Since the extracted candidates are not precise representations of the actual lesions, a region growing step is also applied to them. 
While the original paper \cite{spencer} is written to detect MAs on fluorescein angiographic images, our implementation is based on the modified version published by Fleming et al. \cite{fleming}.

\subsection{Circular Hough-transformation \cite{hough}}
\label{sec:hough}

Following the idea presented in \cite{hough}, we established an approach based on the detection of small circular spots in the image. Candidates are obtained by detecting circles on the images using circular Hough transformation. With this technique, a set of circular objects can be extracted from the image.

\subsection{Zhang et al. \cite{zhang}}
\label{sec:zhang}

In order to extract candidates, this method constructs a maximal correlation response image for the input retinal image. This is accomplished by considering the maximal correlation coefficient with five Gaussian masks with different standard deviations for each pixel. The maximal correlation response image is thresholded with a fixed threshold value to obtain the candidates. Vessel detection and region growing is applied to reduce the number of candidates, and to determine their precise size, respectively.

\subsection{Lazar et al. \cite{lazar}}
\label{sec:ridge}

Pixel-wise cross-section profiles with multiple orientations are used to construct a multi-directional height map. This map assigns a set of height values that describe the distinction of the pixel from its surroundings in a particular direction. In a modified multilevel attribute opening step, a score map is constructed from which the MAs are extracted by thresholding.

\begin{table}
\centering
\caption{Summary of the key differences of the candidate extractors. The sensitivity and average number of false positives per image (FP / I) is measured on the ROC training database with default parameter settings.}
\label{tab:ces}  
\begin{tabular}{|l|l|c|c|c|c|}
\hline
Algorithm & Method & Sensitivity & FP / I\\
\hline
Walter & diameter closing & 36\% & 154.42\\
\hline
Spencer & top-hat transformation & 12\% & 20.3\\
\hline
Hough & circular Hough-transformation & 28\% & 505.85\\
\hline
Zhang & matching multiple Gaussian masks & 33\% & 328.3\\
\hline
Lazar & cross-section profile analysis & 48\% & 73.94\\
\hline
\end{tabular}
\end{table}

\section{Ensemble creation}
\label{sec:com}

In this section, we describe our ensemble creation approach. In our framework, an ensemble $E$ is a set of $\langle$preprocessing method, candidate extractor$\rangle$ or shortly $\langle PP, CE \rangle$ pairs. The meaning of a $\langle$preprocessing method, candidate extractor$\rangle$ pair is that first we apply the preprocessing method to the input image and then we apply the candidate extractor to this result. That is, such a pair will extract a set of candidates $H_{E}$ from the original image.
If an ensemble $E$ contain more $\langle$preprocessing method, candidate extractor$\rangle$ pairs, their outputs are fused in the following way: for each candidate $c$, all such candidates of the other participants are collected, whose euclidean distance $d$ is smaller than a predefined constant $r \in \mathbb{R}$ from $c$. Let $I_{c}$ denote that the set of these points collected for a candidate $c$. Then, the centroid calculated from $I_{c}$ is put into $H_{E}$.

Ensemble creation is a process where all ensembles $E$ from an ensemble pool $\cal{E}$ is evaluated and the best performing one $E_{best} \in \cal{E}$ regarding an evaluation function on a training set is selected. To evaluate an ensemble $E$, its output candidate set $H_{E}$ must be compared to the ground truth in the following way: if for a $c \in H_{E}$ exists a point in the ground truth, whose euclidean distance $d$ from $c$ is smaller than a predefined constant $r \in \mathbb{R}$, then $c$ is considered as a true positive. Otherwise, $c$ is false positive, while each ground truth point is a false negative which does not have a close candidate from $H_{E}$.

The selection of the optimal ensemble $E_{best}$ would require each possible $\langle$preprocessing method, candidate extractor$\rangle$ ensembles to be evaluated to find the optimal one. However, currently we consider $M = N = 5$ preprocessing methods and candidate extractors in our experiments. That is, we have 25 $\langle$preprocessing method, candidate extractor$\rangle$ pairs with $2^{25}$ number of possible combinations to form the ensemble. It would be very resource-demanding to evaluate such a large number of combinations, so we used simulated annealing \cite{sa} as a search algorithm to find the final ensemble, which is proven to be effective in such large search spaces. However, we describe the selection procedure as an exhaustive search in the latter parts, since it is better to evaluate all configurations if enough resources are available, and several other choices of search algorithms are possible.

\begin{figure}[htb]
	\centering
		\includegraphics[width=0.8\linewidth]{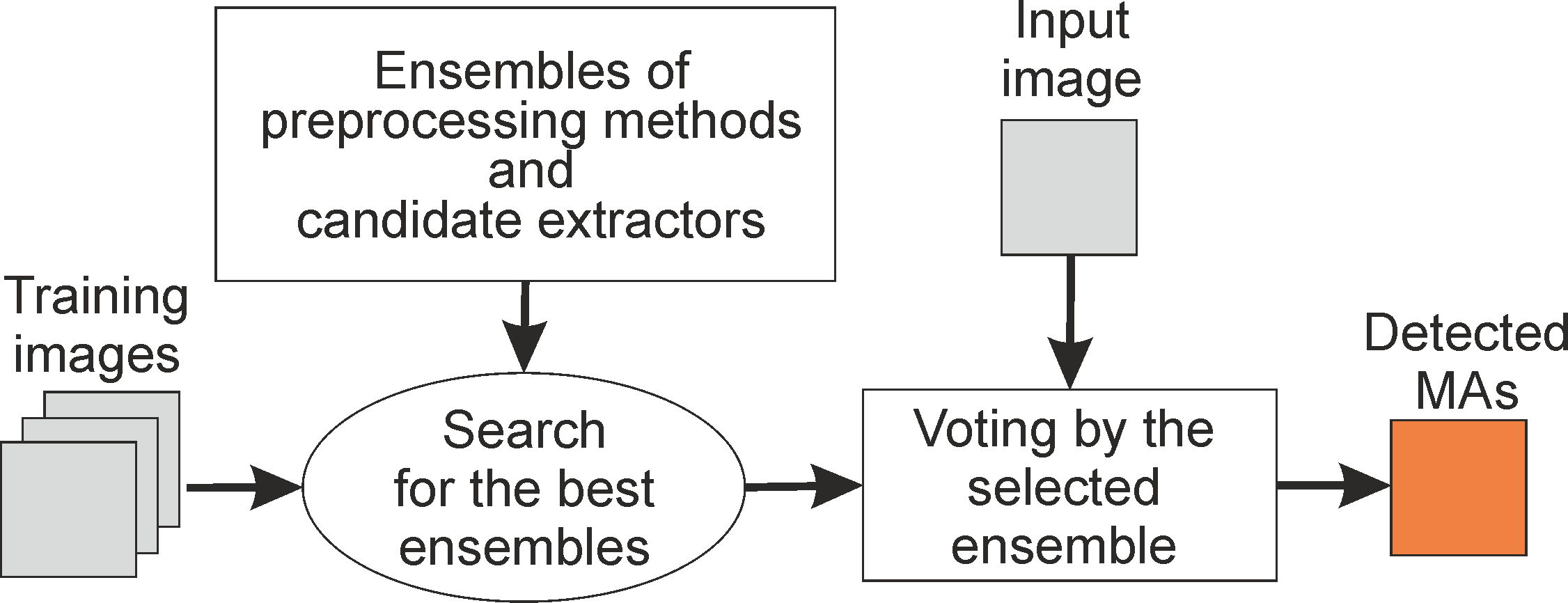}
	\caption{Flow chart of the ensemble-based framework.}
	\label{fig:flow_ensemble}
\end{figure}

As an energy function, we used the competition performance metric CPM \cite{roc}, which is defined as the average sensitivity level at seven predefined false positive per image rate ($1/8, 1/4, 1/2, 1, 2, 4, 8$) \cite{roc}. The process of ensemble creation is also shown in Figure \ref{fig:flow_ensemble}.


The ensemble creation part results in a set of $\langle$preprocessing method, candidate extractor$\rangle$ pairs. This ensemble $E_{best}$ then can be used to detect MAs on unknown images. The final ensemble is applied in real detection in the same way as in the training phase. Namely, the final MAs are detected by the fusion of the MA candidates of the individual pairs building up the ensemble $E_{best}$. Similarly, for every detected MA we will have a confidence value as described above. Thus, for the final decision on the presence of MAs, the output MA set needs to be thresholded according to the assigned confidence values. The choice of the threshold value is discussed in section \ref{sec:dis} in detail.


%

The proposed ensemble creation method can be summarized through the following steps:


\medskip
\hrule
\smallskip
\textbf{Algorithm 1}: Selection of the optimal combination of preprocessing methods and candidate extractors.
\begin{algorithmic}[1]
\algsetup{linenodelimiter=.}
	\hrule
	\medskip
	\STATE $\cal{E}$ $ \leftarrow P\left(PP_{i} \times CE_{j}\right),\,i = 1,\,\dots,\,M,\,j = 1,\,\dots,\,N$ 
	\STATE $CPM_{best} \leftarrow 0$
	\STATE $E_{best} \leftarrow NULL$
	\FORALL{$E \in \cal{E}$}
		\STATE $H_{E} \leftarrow \emptyset$
		\FORALL{$p \in E$}
			\FORALL{MA candidate $c$ detected by $p$}
				\STATE $I_{c} \leftarrow \{ c' |  c'$ is a MA candidate found by a $p' \in E,$ with $p \neq p'$ and $ d\left(c,c'\right) < r\} \cup \{c\}$
				\STATE $confidence\left(c\right) = \dfrac{\left| I_{c} \right|}{\left| E \right|},$
				\STATE $H_{E} \leftarrow H_{E} \cup centroid\left(I_{c}\right)$
			\ENDFOR
		\ENDFOR
		\IF{$CPM\left(H_{E}\right) > CPM_{best}$}
			\STATE $CPM_{best} \leftarrow CPM\left(H_{E}\right)$
			\STATE $E_{best} \leftarrow E$	 
		\ENDIF
	\ENDFOR
	\RETURN $E_{best}$ 
\end{algorithmic}
\medskip
\hrule
\bigskip

\section{Methodology}
\label{sec:methodology}

We have evaluated the proposed approach for both MA detection and DR grading. In this section, we present the evaluation methodology we used in each case.

\subsection{MA detection}
\label{sec:mad}

We have evaluated the MA detection capabilities of the proposed method in the ROC  competition for MA detectors \cite{roc}, as well as on a publicly available \cite{diaretdb1} and a private database. In this section, we provide a brief overview on these databases and on the methodology we used for the evaluation of MA detection performance of the proposed approach. 

\subsubsection{Retinopathy Online Challenge (ROC) \cite{roc}}
\label{sec:comp}

ROC is a worldwide competition dedicated to measure the accuracy of microaneurysm detectors. The ROC database consists of 50 training and 50 test images with different resolutions ($768 \times 576$, $1058 \times 1061$ and $1389 \times 1383$), $45^\circ$ FOV and JPEG compression. The average number of MAs for the training and test sets are 6.72 and 6.86, respectively. There are 13 and 10 images of the training and test sets, where no MAs are marked by the experts.

\subsubsection{DiaretDB1 2.1 database \cite{diaretdb1}}
 
The DiaretDB1 2.1 database contains 28 losslessly compressed training and 61 test  images with a $1500 \times 1152$ resolution and $50^\circ$ FOV. The average number of MAs for the training and test sets are 4.34 and 3.91, respectively. There are 15 and 39 images of the training and test sets, where no MAs are marked by the experts.

\subsubsection{Private database provided by Moorfields Eye Hospital, UK}

This database consists of 60 losslessly compressed images with a resolution $3072 \times 2048$ and $45^\circ$ FOV. The average number of MAs for the training and test sets are 8.67 and 8.87, respectively. There are 10 and 8 images of the training and test sets, respectively, where no MAs are marked by the experts.

\subsubsection{Testing} 

For each database, we provide the Free-response Receiver Operating Characteristic (FROC) curves \cite{froc}, which plots the sensitivity against the average number of false positives per image. To measure the sensitivity at different average false positive per image levels, we thresholded the output set of the MA detector based on the confidence values assigned to each candidate. For the ROC dataset, we also provide the current ranking of the competition along with the CPM values (see section \ref{sec:con} for details) that serves as the basis for the ranking. In addition, we also calculated a partial AUC of the algorithms in the same range (between $1/8$ and $8$) by normalizing the average false positive per image figure by dividing with the maximum ($8$) and applying trapezoidal integration. The empirical AUC calculated this way is likely to underestimate the true AUC. However, the uncertainty for the partial AUCs may be quite high due to the low number of images. 

\subsection{DR grading}
\label{sec:gra}

We have also evaluated our ensemble-based approach to see its grading performance to recognize DR. For this aim, we determined the image-level classification rate of the ensemble on the Messidor\footnote{Kindly provided by the Messidor program partners (see http://messidor.crihan.fr).} dataset containing 1200 images. That is, the presence of any MA means that the image contains signs of DR, while the absence of MAs indicates a healthy case. In other words, a pure yes/no decision of the system has been tested.

\subsubsection{Ensemble creation}
\label{sec:tra}

As there is no training set provided for the Messidor database, we used an independent dataset (the ROC dataset) to train our algorithm. Note that, this is quite a strong handicap in comparison with the usual approach to train on a part of the same database. However, we feel that in this way we can get much closer to measure up the true performance of our system under real circumstances.

\subsubsection{Testing}
\label{sec:tes}

We used the publicly available Messidor database for testing. This database consists of 1200 losslessly compressed images with $45^{\circ}$ FOV and different resolutions ($440 \times 960$, $2240 \times 1488$ and $2304 \times 1536$). For each image, a grading score ranging from R0 to R3 is provided. These grades correspond to the following clinical conditions: a patient with an R0 grade has no DR. R1 and R2 are mild and severe cases of non-proliferative retinopathy, respectively. Finally, R3 is the most serious condition (proliferative retinopathy). The grading is based on the appearance of MAs, haemorrhages and neovascularization. The proportion of the images in the Messidor dataset: 540 R0 (46\%), 153 R1 (12.75\%), 247 R2 (20.58\%) and 260 R3 (21.67\%). 

In our evaluation, we classified the retinal images whether they contain signs of DR (R1, R2, R3) or not (R0). The MA detector classifies an image as diseased if at least one MA was detected, and healthy otherwise. We measured the sensitivity, specificity and accuracy of the detector at different levels by thresholding the confidence values assigned to the MA candidates as described in section \ref{sec:com} using the following formulas:
\[
	sensitivity = \dfrac{tp}{tp+fn},
\]
\[
	specificity = \dfrac{tn}{tn+fp},
\]
and
\[
	accuracy = \dfrac{tp + tn}{tp+fn+tn+fp}.
\]
We also measured that the percentage of correctly recognized cases for each grade. 
We provided a fitted Receiver Operating Characteristic (ROC) curve along with the empirical and fitted AUC for the proposed method on the Messidor database. For curve fitting, we used JROCFIT \cite{jrocfit}.

\section{Results}
\label{sec:res}

In this section, we present our experimental results for both MA detection and DR grading.

\subsection{MA detection}

In Table \ref{tab:pairs}, we exhibit the $\langle$preprocessing method, candidate extractor$\rangle$ pairs included in the selected ensembles for the three datasets, respectively. The rows of the table show the preprocessing methods from section \ref{sec:pre}, while the columns label the candidate extractor algorithms listed in section \ref{sec:ces}. 
\begin{table}
\centering
\caption{$\langle$Preprocessing method, candidate extractor$\rangle$ pairs selected as members of the ensemble for the three dataset. R, D, M denote whether the pair is selected for the ROC, Diaret2.1, or the Moorfields dataset, respectively.}
\label{tab:pairs}  
\begin{tabular}{|l|c|c|c|c|c|}
\hline
& Walter & Spencer & Hough & Lazar & Zhang\\
\hline
Walter-Klein & & M & & & R\\
\hline
CLAHE & R, D & M & & R & D\\
\hline
Vessel Removal & D & & & R, D, M & R, D\\
\hline
Illumination eq. & & & & R, M &\\
\hline
No preprocessing & R &  & M & R, D & R\\
\hline
\end{tabular}
\end{table}

\begin{figure}[htb]
	\centering
		\includegraphics[width=0.8\linewidth]{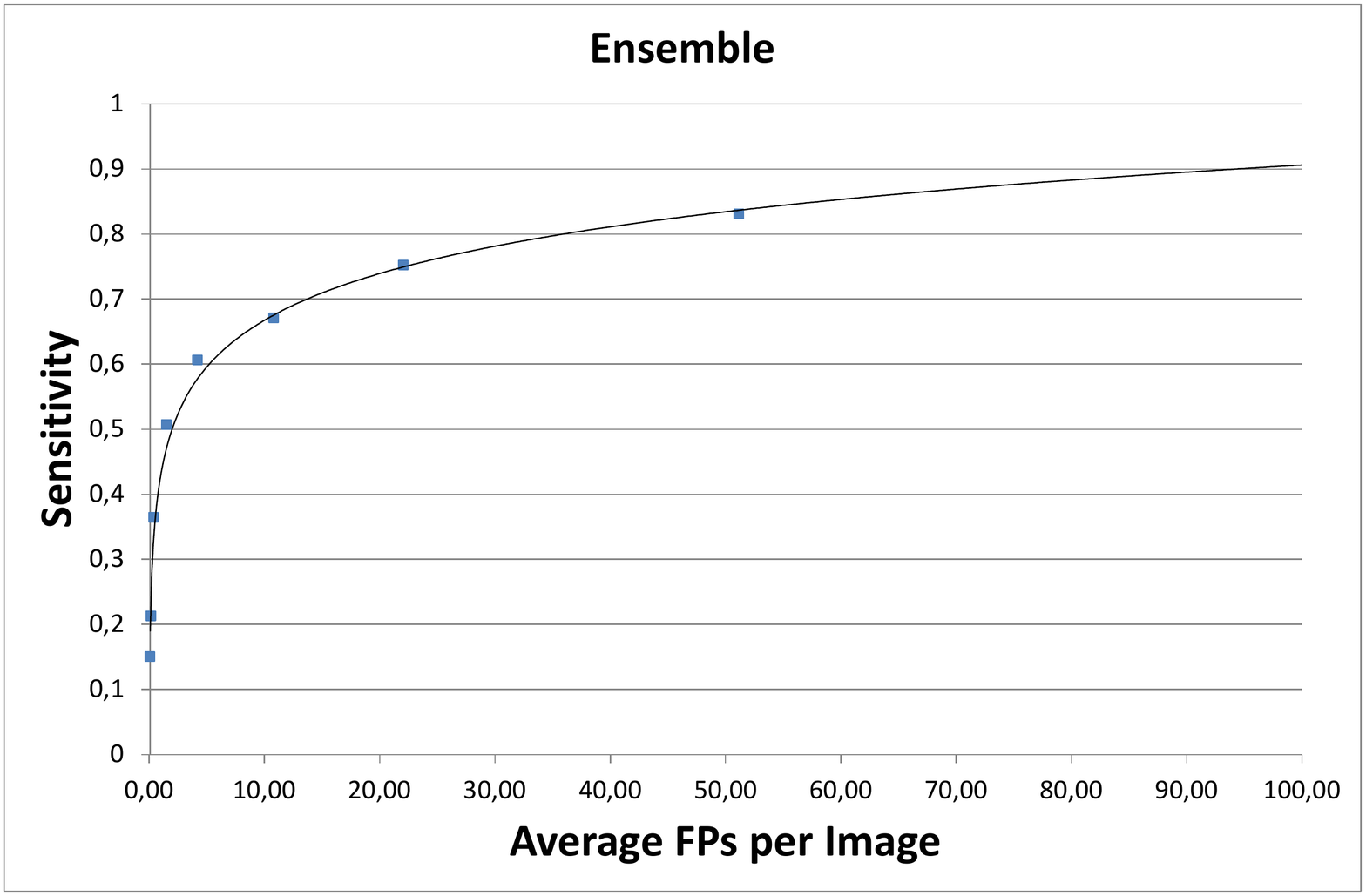}
	\caption{FROC curve of the ensemble on the ROC dataset.}
	\label{fig:roc_ens}
\end{figure}

Table \ref{tab:res} contains the ranked quantitative results of the participants at the ROC competition, with the proposed ensemble (DRSCREEN) highlighted as the current leader. The performance of the ensemble is also shown in Figure \ref{fig:roc_ens} in terms of a FROC curve. As we can see from Table \ref{tab:res}, the proposed ensemble earned both a higher CPM score and a higher partial AUC than the individual algorithms. 


\begin{table}[ht]
\centering
\caption{Quantitative results of the ROC competition. For each participating team, the competition performance metric and the partial AUC are presented.}
\label{tab:res}
\begin{tabular}{|c|c|c|}
\hline
Team & CPM & AUC\\
\hline
\textbf{DRSCREEN} & \textbf{0.434} & \textbf{0.551}\\
\hline
 Niemeijer et al. &   0.395 & 0.469\\
\hline
 LaTIM &  0.381 & 0.489\\
\hline
 ISMV &  0.375 & 0.435\\
\hline
 OKmedical II &  0.369 & 0.465\\
\hline
 OKmedical &   0.357 & 0.430\\
\hline
Lazar et al. &  0.355 & 0.449\\
\hline
 GIB &   0.322 & 0.399\\
\hline
 Fujita &    0.310 & 0.378\\
\hline
 IRIA &   0.264 & 0.368\\
\hline
 Waikato &  0.206 & 0.273\\
\hline
\end{tabular}  
\end{table}

\begin{figure}[htb]
	\centering
		\includegraphics[width=0.8\linewidth]{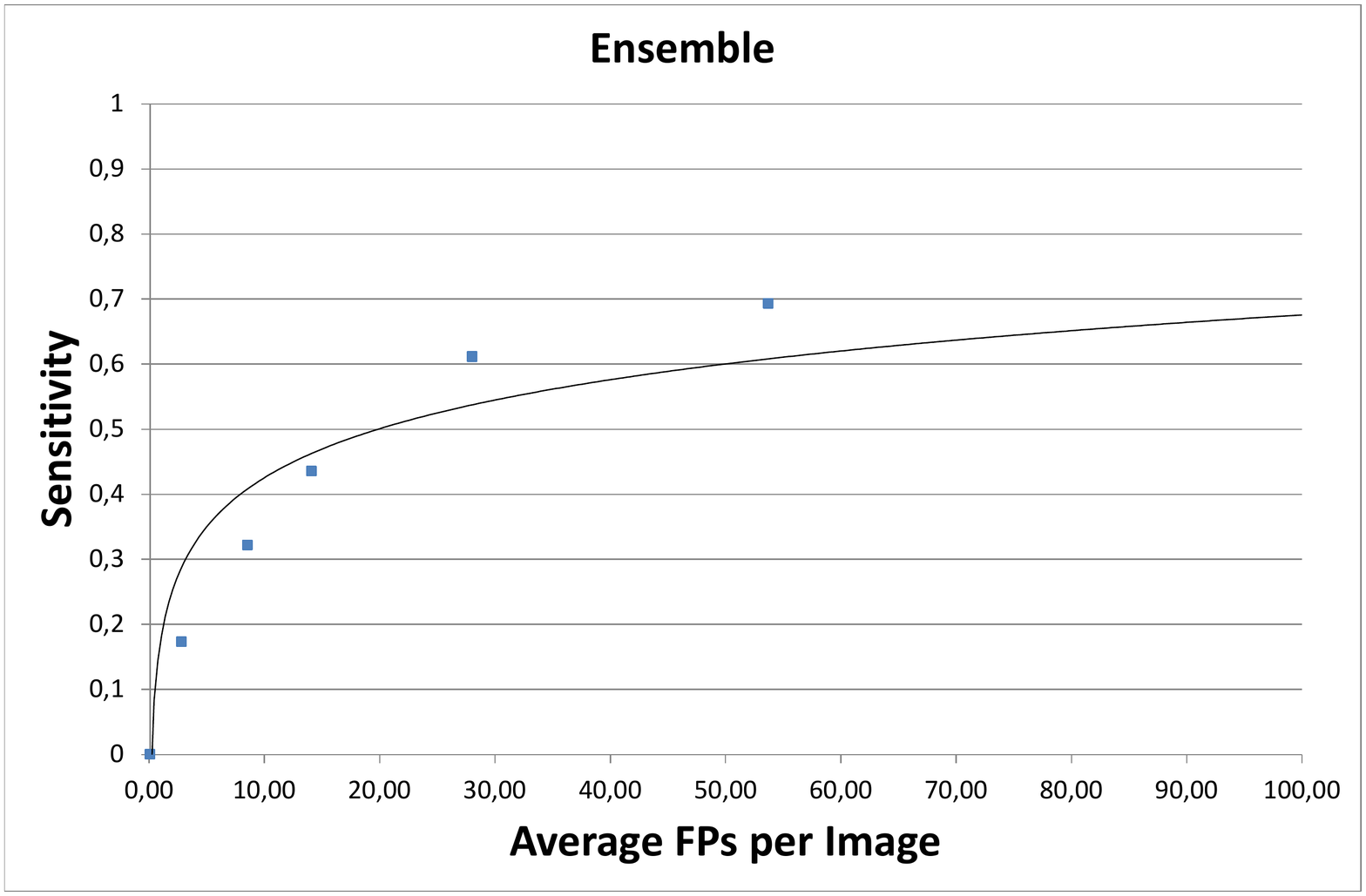}
	\caption{FROC curve of the ensemble on the DiaretDB2.1 dataset.}
	\label{fig:diaret_ens}
\end{figure}

\begin{figure}[htb]
	\centering
		\includegraphics[width=0.8\linewidth]{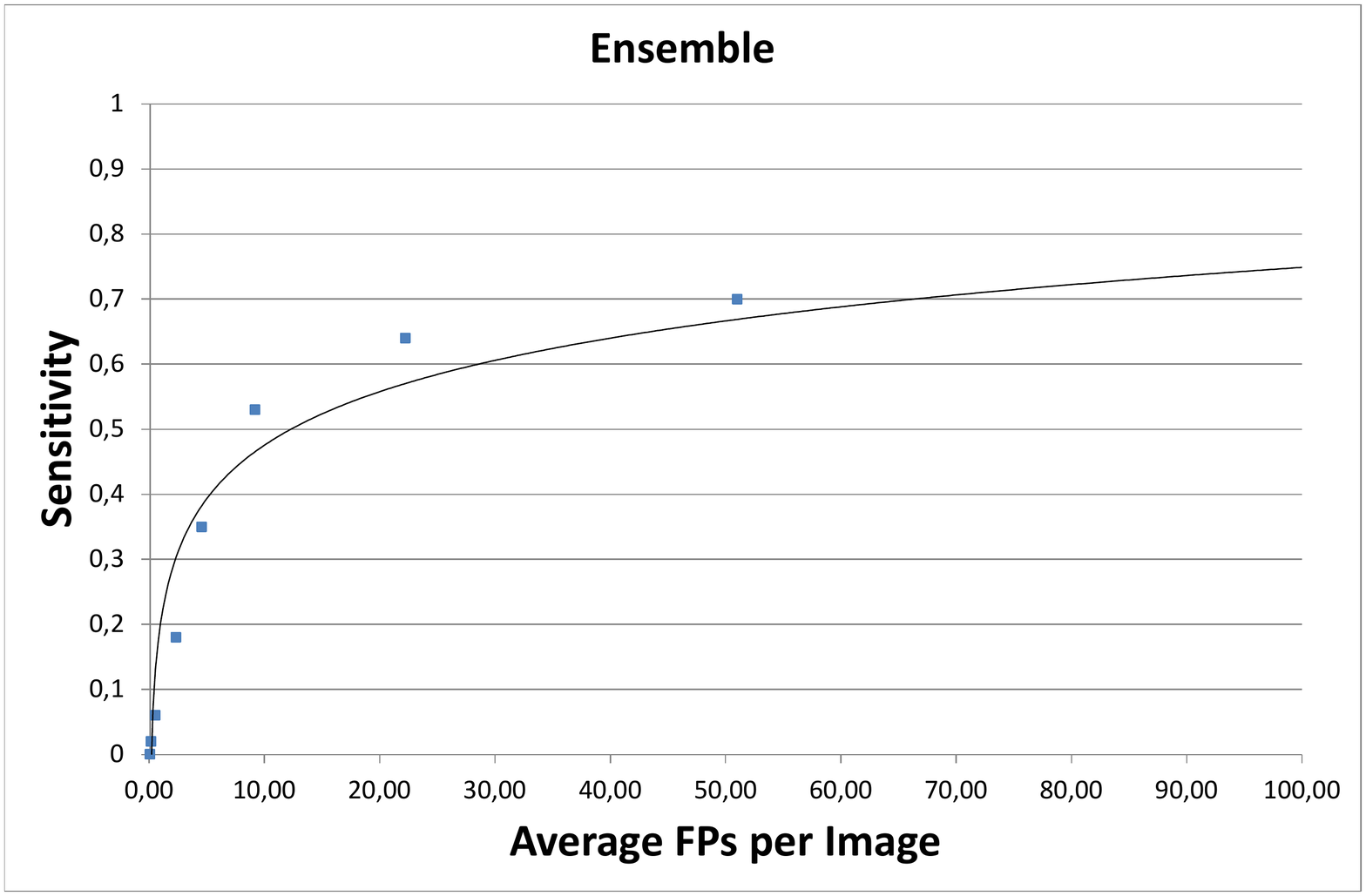}
	\caption{FROC curve of the ensemble on the Moorfields dataset.}
	\label{fig:moorefields_ens}
\end{figure}

The FROC curves of the ensemble for the DiaretDB1 v2.1 and for the Moorfields database is shown in Figures \ref{fig:diaret_ens} and \ref{fig:moorefields_ens}, respectively. To the best of our knowledge, no corresponding quantitative results have been published for these databases yet. Thus, we disclose the results of the ensemble-based method only.

\subsection{DR grading}   

\begin{figure}[htb]
	\centering
		\includegraphics[width=0.8\linewidth]{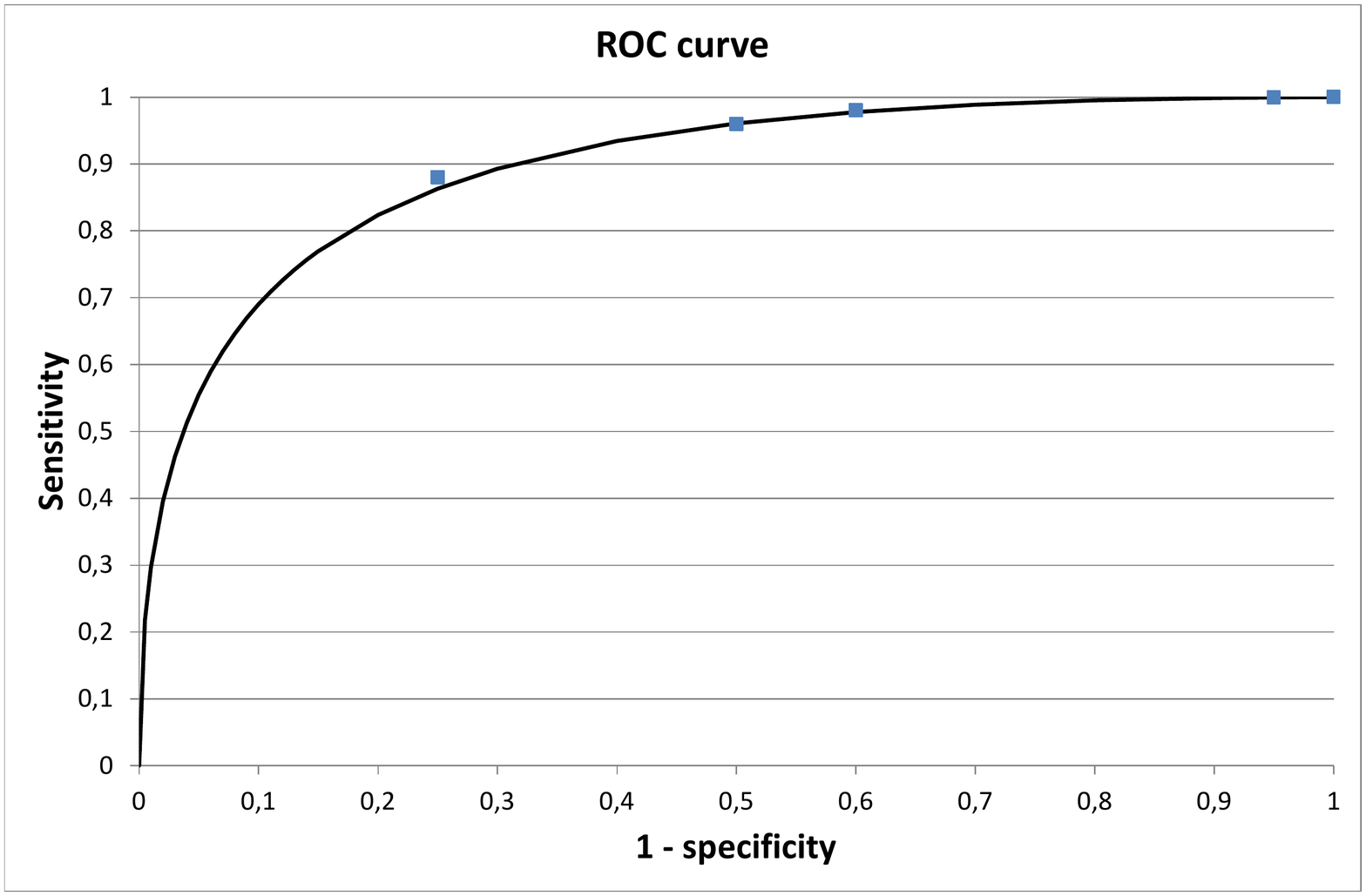}
	\caption{ROC curve of the ensemble on the Messidor dataset.}
	\label{fig:messidor}
\end{figure}

In Table \ref{tab:messidor}, we provide the sensitivity, specificity and accuracy measures of our detector corresponding to different threshold values, respectively. The fitted ROC curve of the detector can be seen in Figure \ref{fig:messidor}. The empirical area under curve (AUC) is 0.875, while the AUC for the fitted curve is $0.90\pm0.01$. Table \ref{tab:messidor} also contains the percentage of the correctly recognized cases for each class.  
   
\begin{table}
	\caption{Results on the Messidor dataset. For each threshold, sensitivity, specificity, accuracy and the percentage of correctly recognized cases for each grade are presented.}
	\label{tab:messidor}
	\begin{center}
		\begin{tabular}{|c|c|c|c|c|c|c|c|}
		\hline
		Threshold& 0.4 & 0.5 & 0.6 & 0.7 & 0.8 & 0.9 & 1.0\\
		\hline
		Sensitivity & 1	& 1 & 1 & 0.99 & 0.96 &	0.76 & 0.31\\
		\hline
		Specificity & 0 & 0.01 &	0.03 &	0.14 &	0.51 &	0.88 &	0.98\\
		\hline
		Accuracy & 0.53 &	0.54 &	0.55 &	0.59 &	0.75 &	0.82 &	0.62\\
		\hline
		R0 & 0.00 &	0.01 &	0.03 &	0.14 &	0.51 & 0.88 & 0.98\\
		\hline
		R1 & 1.00 & 1.00 & 1.00 & 0.97 & 0.92 & 0.60 & 0.18\\
		\hline
		R2 & 1.00 & 1.00 & 1.00 & 1.00 & 0.96 & 0.72 & 0.29\\
		\hline
		R3 & 1.00 & 1.00 & 1.00 & 1.00 & 0.98 &	0.92 & 0.42\\
		\hline
		\end{tabular}
	\end{center}

\end{table}

%
%
%

\section{Discussion}
\label{sec:dis}

\begin{figure}[htb]
	\centering
	\subfigure[Original]{
		\includegraphics[width=0.15\linewidth]{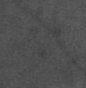}
	}
	\quad
	\subfigure[CLAHE]{
		\includegraphics[width=0.15\linewidth]{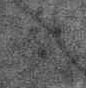}
	}
	\quad
	\subfigure[Vessel removal]{
		\includegraphics[width=0.15\linewidth]{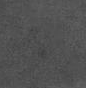}
	}
	\caption{The effect of different preprocessing methods where MAs are hard to detect.}
	\label{fig:difficulty}
\end{figure}

A strong point of the proposed method is that it performs well under difficult circumstances. Figure \ref{fig:difficulty} shows an example image where the application of CLAHE made it easier to distinguish the MAs from their background. However, the use of the vessel removal and inpainting preprocessing method caused the missing of a true MA, while the detection of the remaining MA is easier in the absence of thin retinal vessels. Thus, using different preprocessing methods with candidate extractors creates diversity among the members of the ensemble, which is desired for systems using multiple estimators \cite{kuncheva}. This diversity ensures the suppression of false detections, since diverse detectors tend to make different mistakes. Thus, the false detections are likely to receive lower confidence values in the voting procedure.   

Our experimental results show that the proposed ensemble-based MA detector outperforms the current individual approaches in MA detection. It has been also proven that the framework has high flexibility for different datasets. As it can be seen in Table \ref{tab:pairs}, the ensemble members may vary, which suggests relatively high variance among databases in this field. Despite this variability, the performance of the ensemble still remained stable. In \cite{roc}, the authors measured a human expert average false positive rate at the ROC dataset against the consensus of three human experts. This level is approximately 1 FP per image \cite{roc} for the ROC database, on which level our ensemble achieved the best score in the competition. Thus, we can recommend to use this level for thresholding at the ensemble creation phase and use it for detecting MAs on unknown images.

As for DR grading, our ensemble also performed well. It is also important to see how the different classes (R0, R1, R2, R3) are recognized at different levels. As it can be desired, the severity of DR affects the performance of our detector. At each threshold level, where the sensitivity is less than 1.0, the more severe case recognized with higher probability.  

The selection of the appropriate threshold is also an important issue for our detector to provide sufficient sensitivity and specificity rate. In \cite{early}, the authors suggest that sensitivity is more important for a screening system than specificity. In opposition, the British Diabetic Association (BDA) recommends 80\% sensitivity and 95\% specificity for DR screening \cite{bda}. In Table \ref{tab:messidor}, we can see that the most accurate result is achieved with the threshold value 0.9. By applying the first idea, we might consider the results corresponding to the threshold value 0.8 as the best in our experiment, where 96\% sensitivity and 51\% specificity are achieved. That is, we recognized almost all of the cases where DR is present, and half of the healthy ones. The closest to the second recommendation is the performance achieved at the 0.9 level: 76\% sensitivity and 88\% specificity.


It is difficult to compare our method to other screening systems. First of all, to the best of our knowledge, no other results reported for the complete Messidor database. Other screening systems are tested on private images. Unfortunately, the proportion of non-DR/DR cases are varying in these experiments. Abramoff et al.~\cite{early} reported 0.86 AUC on a population where 4.96\% of the cases had at least minimum signs of DR. The databases on which Agurto et al.~\cite{Agurto} tested, 74.43\% and 76.26\% cases contained signs of DR and they achieved 0.81 and 0.89 AUCs, respectively. The closest to match the requirements of BDA is the system of Jelinek et al.~\cite{Jelinek2006} with a 85\% sensitivity and 90\% specificity, where approximately 30\% of patients had DR. Similar proportion (35.88\%) of patients having DR are reported by Fleming et al.~\cite{largescale} in their automatic screening system. 

Despite the promising results, our system still misclassifies some stage where serious case of DR is present. To improve grading performance, we must take into account the presence or absence of more DR-specific lesions (e.g. exudates), which are essential in a clinical setting. However, our MA detector can serve as a main component of such a system.       


\section{Conclusion}
\label{sec:con}

In this paper, we have proposed an ensemble-based microaneurysm detector which has proved its high efficiency in an open online challenge with its first position. Our novel framework relies on a set of $\langle$preprocessing method, candidate extractor$\rangle$ pairs, from which a search algorithm selects an optimal combination. Since our approach is modular, we can expect further improvements by adding more preprocessing methods and candidate extractors. We have also evaluated the grading performance of this detector in the 1200 images of the Messidor database. We have achieved a $0.90\pm0.01$ AUC value, which is competitive with the previously reported results on other databases. The grading results presented in this paper are already promising. However, a proper screening system should contain other components, which is expected to increase the performance of this approach, as well.  

\section*{Acknowledgement}
This work was supported in part by the J\'anos Bolyai grant of the Hungarian Academy of Sciences, and by the TECH08-2 project DRSCREEN - Developing a computer based image processing system for diabetic retinopathy screening of the National
Office for Research and Technology of Hungary (contract no.: OM-00194/2008, OM-00195/2008, OM-00196/2008).

\balance

\bibliographystyle{IEEETran} 
\bibliography{refs_rev}

\begin{thebibliography}{10}
\providecommand{\url}[1]{#1}
\csname url@samestyle\endcsname
\providecommand{\newblock}{\relax}
\providecommand{\bibinfo}[2]{#2}
\providecommand{\BIBentrySTDinterwordspacing}{\spaceskip=0pt\relax}
\providecommand{\BIBentryALTinterwordstretchfactor}{4}
\providecommand{\BIBentryALTinterwordspacing}{\spaceskip=\fontdimen2\font plus
\BIBentryALTinterwordstretchfactor\fontdimen3\font minus
  \fontdimen4\font\relax}
\providecommand{\BIBforeignlanguage}[2]{{%
\expandafter\ifx\csname l@#1\endcsname\relax
\typeout{** WARNING: IEEEtran.bst: No hyphenation pattern has been}%
\typeout{** loaded for the language `#1'. Using the pattern for}%
\typeout{** the default language instead.}%
\else
\language=\csname l@#1\endcsname
\fi
#2}}
\providecommand{\BIBdecl}{\relax}
\BIBdecl

\bibitem{system}
M.~Abramoff, M.~Niemeijer, M.~Suttorp-Schulten, M.~A. Viergever, S.~R. Russel,
  and B.~van Ginneken, ``Evaluation of a system for automatic detection of
  diabetic retinopathy from color fundus photographs in a large population of
  patients with diabetes,'' \emph{Diabetes Care}, vol.~31, pp. 193--198, 2008.

\bibitem{largescale}
A.~D. Fleming, K.~A. Goatman, S.~Philip, G.~J. Prescott, P.~F. Sharp, and J.~A.
  Olson, ``Automated grading for diabetic retinopathy: a large-scale audit
  using arbitration by clinical experts,'' \emph{British Journal of
  Ophthalmology}, vol.~94, no.~12, pp. 1606--1610, 2010.

\bibitem{Jelinek2006}
H.~J. Jelinek, M.~J. Cree, D.~Worsley, A.~Luckie, and P.~Nixon, ``An automated
  microaneurysm detector as a tool for identification of diabetic retinopathy
  in rural optometric practice.'' \emph{Clin Exp Optom}, vol.~89, no.~5, pp.
  299--305, 2006.

\bibitem{early}
M.~Abramoff, J.~Reinhardt, S.~Russell, J.~Folk, V.~Mahajan, M.~Niemeijer, and
  G.~Quellec, ``Automated early detection of diabetic retinopathy,''
  \emph{Ophthalmology}, vol. 117, no.~6, pp. 1147--1154, 2010.

\bibitem{combining}
M.~Niemeijer, M.~Loog, M.~D. Abramoff, M.~A. Viergever, M.~Prokop, and B.~van
  Ginneken, ``On combining computer-aided detection systems,'' \emph{IEEE
  Transactions on Medical Imaging}, vol.~30, pp. 215 -- 223, 2011.

\bibitem{sofa}
B.~Antal, I.~Lazar, A.~Hajdu, Z.~Torok, A.~Csutak, and T.~Peto, ``A multi-level
  ensemble-based system for detecting microaneurysms in fundus images,'' in
  \emph{4th IEEE International Workshop on Soft Computing Applications (SOFA
  2010)}, 2010, pp. 137--142.

\bibitem{pr}
B.~Antal and A.~Hajdu, ``Improving microaneurysm detection using an optimally
  selected subset of candidate extractors and preprocessing methods,''
  \emph{Pattern Recognition}, vol.~45, no.~1, pp. 264 -- 270, 2012.

\bibitem{youssif}
A.~A.~A. Youssif, A.~Z. Ghalwash, and A.~S. Ghoneim, ``Comparative study of
  contrast enhancement and illumination equalization methods for retinal
  vasculature segmentation,'' \emph{Proc. Cairo International Biomedical
  Engineering Conferemce}, 2006.

\bibitem{ce}
T.~Walter and J.~Klein, ``Automatic detection of microaneyrysms in color fundus
  images of the human retina by means of the bounding box closing,''
  \emph{Lecture Notes in Computer Science}, vol. 2526, pp. 210--220, 2002.

\bibitem{clahe}
K.~Zuiderveld, ``Contrast limited adaptive histogram equalization,''
  \emph{Graphics gems}, vol.~IV, pp. 474--485, 1994.

\bibitem{cvpr}
S.~Ravishankar, A.~Jain, and A.~Mittal, ``Automated feature extraction for
  early detection of diabetic retinopathy in fundus images,'' in \emph{Computer
  Vision and Pattern Recognition}, 2009, pp. 210--217.

\bibitem{inpaint}
A.~Criminisi, P.~Perez, and K.~Toyama, ``Object removal by exemplar-based
  inpainting,'' in \emph{Computer Vision and Pattern Recognition}, vol.~2,
  2003, pp. II--721 -- II--728.

\bibitem{roc}
M.~Niemeijer, B.~van Ginneken, M.~Cree, A.~Mizutani, G.~Quellec, C.~Sanchez,
  B.~Zhang, R.~Hornero, M.~Lamard, C.~Muramatsu, X.~Wu, G.~Cazuguel, J.~You,
  A.~Mayo, Q.~Li, Y.~Hatanaka, B.~Cochener, C.~Roux, F.~Karray, M.~Garcia,
  H.~Fujita, and M.~Abramoff, ``Retinopathy online challenge: Automatic
  detection of microaneurysms in digital color fundus photographs,'' \emph{IEEE
  Transactions on Medical Imaging}, vol.~29, no.~1, pp. 185--195, 2010.

\bibitem{walter}
T.~Walter, P.~Massin, A.~Arginay, R.~Ordonez, C.~Jeulin, and J.~C. Klein,
  ``Automatic detection of microaneurysms in color fundus images,''
  \emph{Medical Image Analysis}, vol.~11, pp. 555--566, 2007.

\bibitem{spencer}
T.~Spencer, J.~A. Olson, K.~C. McHardy, P.~F. Sharp, and J.~V. Forrester, ``An
  image-processing strategy for the segmentation and quantification of
  microaneurysms in fluorescein angiograms of the ocular fundus,''
  \emph{Computers and Biomedical Research}, vol.~29, pp. 284--302, 1996.

\bibitem{fleming}
A.~D. Fleming, S.~Philip, and K.~A. Goatman, ``Automated microaneurysm
  detection using local contrast normalization and local vessel detection,''
  \emph{IEEE Transactions on Medical Imaging}, vol. 25(9), pp. 1223--1232,
  2006.

\bibitem{hough}
S.~Abdelazeem, ``Microaneurysm detection using vessels removal and circular
  hough transform,'' \emph{Proceedings of the Nineteenth National Radio Science
  Conference}, pp. 421 -- 426, 2002.

\bibitem{zhang}
B.~Zhang, X.~Wu, J.~You, Q.~Li, and F.~Karray, ``Detection of microaneurysms
  using multi-scale correlation coefficients,'' \emph{Pattern Recogn.},
  vol.~43, no.~6, pp. 2237--2248, 2010.

\bibitem{lazar}
I.~Lazar and A.~Hajdu, ``Microaneurysm detection in retinal images using a
  rotating cross-section based model,'' in \emph{2011 IEEE International
  Symposium on Biomedical Imaging}, 2011, pp. 1405 --1409.

\bibitem{sa}
S.~Kirkpatrick, C.~D. Gelatt, and M.~P. Vecchi, ``Optimization by simulated
  annealing,'' \emph{Science}, vol. 220, pp. 671--680, 1983.

\bibitem{diaretdb1}
T.~Kauppi, V.~Kalesnykiene, J.-K. Kämäräinen, L.~Lensu, I.~Sorri, A.~Raninen,
  R.~Voutilainen, H.~Uusitalo, H.~Kälviäinen, and J.~Pietilä, ``Diaretdb1
  diabetic retinopathy database and evaluation protocol,'' \emph{Proc. of the
  11th Conf. on Medical Image Understanding and Analysis (MIUA2007)}, pp.
  61--65, 2007.

\bibitem{froc}
D.~Chakraborty, ``Clinical relevance of the roc and free-response paradigms for
  comparing imaging system efficacies,'' \emph{Radiation Protection Dosimetry},
  vol. 139, no. 1-3, pp. 37--41, 2010.

\bibitem{jrocfit}
\BIBentryALTinterwordspacing
J.~Eng. Roc analysis: web-based calculator for roc curves. Johns Hopkins
  University, Baltimore. [Online]. Available: \url{http://www.jrocfit.org}
\BIBentrySTDinterwordspacing

\bibitem{kuncheva}
L.~I. Kuncheva, \emph{Combining Pattern Classifiers. Methods and
  Algorithms}.\hskip 1em plus 0.5em minus 0.4em\relax Wiley, 2004.

\bibitem{bda}
B.~D. Association, ``Retinal photography screening for diabetic eye disease,''
  Tech. Rep., 1997.

\bibitem{Agurto}
C.~Agurto, E.~S. Barriga, V.~Murray, S.~Nemeth, R.~Crammer, W.~Bauman,
  G.~Zamora, M.~S. Pattichis, and P.~Soliz, ``Automatic detection of diabetic
  retinopathy and age-related macular degeneration in digital fundus images,''
  \emph{Investigative Ophthalmology \& Visual Science}, vol.~52, no.~8, pp.
  5862--5871, 2011.

\end{thebibliography}

\end{document}